\documentclass[conference]{IEEEtran}

\IEEEoverridecommandlockouts

\usepackage{cite}

\usepackage{amsmath,amssymb,amsfonts}

\usepackage{algorithmic}

\usepackage{graphicx}

\usepackage{textcomp}

\usepackage{xcolor}

\usepackage{acro}

\usepackage{multirow}

\usepackage{booktabs}

\usepackage{comment}

\usepackage{graphicx}  

\usepackage{subfig}    

\usepackage{caption}   

\usepackage{graphicx}

\usepackage{subfig}

\usepackage{graphicx}

\usepackage{float}

\usepackage{caption}

\usepackage{hyperref}

\usepackage{tikz}

\usetikzlibrary{arrows.meta, positioning, calc}

\newcommand*\circled[1]{\tikz[baseline=(char.base)]{
            \node[shape=circle,draw,inner sep=2pt] (char) {#1};}}

\usepackage{pgfplots}

\usepackage{geometry}

\DeclareAcronym{ML}{
  short=ML,
  long=Machine Learning,
}
\DeclareAcronym{SVM}{
  short=SVM,
  long=Support Vector Machine,
}
\DeclareAcronym{CNN}{
  short=CNN,
  long=Convolutional Neural Network,
}
\DeclareAcronym{NIDS}{
  short=NIDS,
  long=Network Intrusion Detection System,
}
\DeclareAcronym{DNN}{
  short=DNN,
  long=Deep Neural Network,
}
\DeclareAcronym{FGSM}{  
  short=FGSM,
  long=Fast Gradient Sign Method,
}
\DeclareAcronym{PGD}{
  short=PGD,
  long=Projected Gradient Descent,
}
\DeclareAcronym{IDS}{
  short=IDS,
  long=Intrusion Detection System,
}

\def\BibTeX{{\rm B\kern-.05em{\sc i\kern-.025em b}\kern-.08em

    T\kern-.1667em\lower.7ex\hbox{E}\kern-.125emX}}

\begin{document}


\title{\Large \textbf{ A No-Defense Defense Against Gradient-Based Adversarial Attacks on ML-NIDS: Is Less More?}\\
\small Authors’ draft for soliciting feedback}




\author{
  \IEEEauthorblockN{Mohamed elShehaby, Ashraf Matrawy}
  \IEEEauthorblockA{\textit{Carleton University}, Ottawa, Canada}
}

\maketitle

\begin{abstract}

Gradient-based adversarial attacks subtly manipulate inputs of Machine Learning (ML) models to induce incorrect predictions. This paper investigates whether careful architectural choices alone can yield an \textit{inherently} robust Deep Neural Network (DNN)-based Network Intrusion Detection Systems (NIDS), without any additional explicit defenses. Through thousands of experiments, around 2200, varying network depth, feature dimensionality, activation functions, and dropout across FGSM, PGD, and BIM attacks, we show that shallower networks, reduced feature sets, and ReLU activation consistently and jointly reduce adversarial vulnerability. Moreover, a simple model following this recipe outperforms deeper, fully-featured adversarially trained models, while maintaining near-perfect clean-traffic detection and lower training times. Nevertheless, while less is more, the selection of the \textit{right less} is what truly matters.


\end{abstract}


\section{Introduction}

Deep Neural Network (DNN)-based Network Intrusion Detection Systems (NIDS) are vulnerable to gradient-based adversarial attacks, carefully crafted perturbations that cause the model to misclassify malicious traffic as benign. These attacks serve as the de facto benchmark for robustness evaluations because they represent a theoretical worst-case adversary. By leveraging the network's loss gradients, they calculate the exact, mathematically optimal path required to cross a decision boundary, rigorously exposing the model's underlying structural limits. Existing defenses, such as adversarial training, are effective but computationally expensive and reactive. This raises a more fundamental Research Question (RQ): \hypertarget{target:RQ}{\textit{Does the model have to be vulnerable in the first place?}}

This paper argues that it does not. Through thousands of experiments varying network depth, feature dimensionality, activation functions, and dropout against numerous gradient-based attacks, we show that deliberate architectural choices alone can yield an \textit{inherently} robust NIDS, which we term a \textit{``no-defense defense''}, requiring no explicit defense mechanism.


The main contributions of this paper are: \textbf{\circled{1}} A large-scale empirical study of how depth, feature dimensionality, activation functions, and dropout affect adversarial robustness in DNN-based NIDS. \textbf{\circled{2}} Evidence that a shallow, reduced-feature, ReLU-based model constitutes an inherent no-defense defense against NIDS' gradient-based adversarial attacks. \textbf{\circled{3}} A demonstration that this recipe outperforms adversarially trained deep models on robustness, clean-traffic performance, and training efficiency. \textbf{\circled{4}} An exploration of the practical relevance of gradient-based attacks in the NIDS domain and the implications for real-world deployment.


\begin{figure}
    \centering
    
\begin{tikzpicture}
\begin{axis}[
    xlabel={},
    ylabel={},
    zlabel={Loss $J$},
    view={130}{35}, 
    colormap/viridis,
    grid=major,
    ticks=none,
    title style={yshift=10pt},
    label style={font=\small},
    zlabel style={rotate=-90, anchor=south},
]
    \addplot3[
        surf,
        domain=-2:2,
        domain y=-2:2,
        opacity=0.35, 
        samples=30,
        shader=interp
    ] {exp(-(x^2 + y^2)) * -6}; 

    \draw[->, ultra thick, red!80!black] (axis cs:0,0,-6) -- (axis cs:1.2,1.2,-1.5) 
        node[midway, sloped, above, color=black, font=\bfseries\tiny, yshift=2pt] {Direction of $\nabla_x J$};

    \addplot3[only marks, mark=*, mark size=3pt, blue] coordinates {(0,0,-6)};
    \node[blue, anchor=north] at (axis cs:0,0,-6.2) {\textbf{$x$}};

    \addplot3[only marks, mark=*, mark size=2pt, orange] coordinates {(0.4,0.4,-5.2)};
    \node[orange, anchor=west, xshift=5pt, font=\scriptsize] at (axis cs:0.4,0.4,-5.2) {$\epsilon=0.1$};

    \addplot3[only marks, mark=*, mark size=2pt, orange] coordinates {(0.8,0.8,-3.5)};
    \node[orange, anchor=west, xshift=5pt, font=\scriptsize] at (axis cs:0.8,0.8,-3.5) {$\epsilon=0.3$};

    \addplot3[only marks, mark=*, mark size=2pt, orange] coordinates {(1.2,1.2,-1.5)};
    \node[orange, anchor=west, xshift=5pt, font=\scriptsize] at (axis cs:1.2,1.2,-1.5) {$\epsilon=0.5$};

\end{axis}
\end{tikzpicture}

\caption{Visualization of the FGSM attack mechanism on a 3D loss landscape. The figure illustrates the shift of the original input $x$ in the direction of the loss gradient $\nabla_{x}J(\theta,x,y)$ across varying perturbation magnitudes $\epsilon$.}
    \label{fig:CF_ADV}
\end{figure}
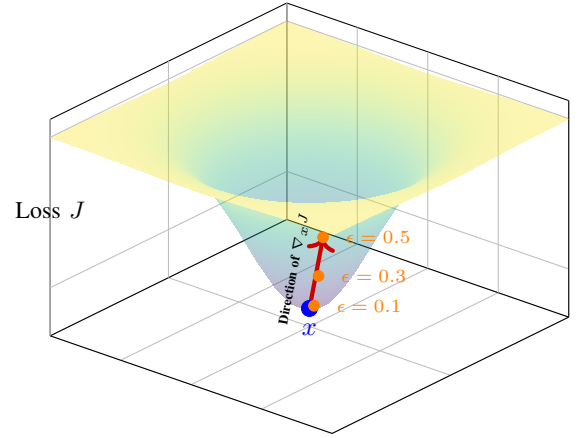


\section{Gradient-Based Adversarial Attacks}

Gradient-based adversarial attacks against ML models involve adding carefully crafted perturbations to input data that are imperceptible to humans but can cause significant misclassification. These attacks were first systematically studied in the computer vision domain.

\subsection{Fast Gradient Sign Method (FGSM)}
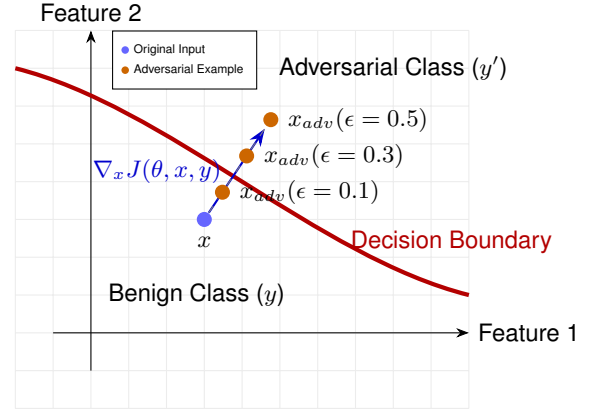
\begin{figure}[]
    \centering
    
\begin{tikzpicture}[
    font=\sffamily\small,
    >=Stealth,
    decision_boundary/.style={ultra thick, color=red!70!black},
    data_point/.style={circle, fill=blue!60, inner sep=2pt},
    adv_point/.style={circle, fill=orange!80!black, inner sep=2pt}
]

    \draw[lightgray!30, thin, step=0.5] (-1,-1) grid (5,4);
    \draw[->] (-0.5,0) -- (5,0) node[right] {Feature 1};
    \draw[->] (0,-0.5) -- (0,4) node[above] {Feature 2};

    \draw[decision_boundary] (-1, 3.5) .. controls (1, 3) and (3, 1) .. (5, 0.5);
    \node[color=red!70!black] at (4.77, 1.2) {Decision Boundary};
    \node at (1.4, 0.5) {Benign Class ($y$)};
    \node at (4, 3.5) {Adversarial Class ($y'$)};

    \node[data_point, label=below:{$x$}] (X) at (1.5, 1.5) {};

    \draw[->, thick, blue!80!black] (X) -- ++(0.8, 1.2) node[midway, left, xshift=-2pt] {$\nabla_x J(\theta, x, y)$};

    \node[adv_point, label=right:{$x_{adv} (\epsilon = 0.1)$}] (E1) at ($(X) + 0.3*(0.8, 1.2)$) {};
    \draw[dashed, gray] (X) -- (E1);

    \node[adv_point, label=right:{$x_{adv} (\epsilon = 0.3)$}] (E2) at ($(X) + 0.7*(0.8, 1.2)$) {};
    \draw[dashed, gray] (E1) -- (E2);

    \node[adv_point, label=right:{$x_{adv} (\epsilon = 0.5)$}] (E3) at ($(X) + 1.1*(0.8, 1.2)$) {};
    \draw[dashed, gray] (E2) -- (E3);

    \matrix [draw, below left, fill=white, nodes={scale=0.5}] at (2.2, 4) {
      \node [data_point, label=right:Original Input] {}; \\
      \node [adv_point, label=right:Adversarial Example] {}; \\
    };

\end{tikzpicture}
\caption{Conceptual 2D representation of a gradient-based evasion attack. As the perturbation magnitude $\epsilon$ increases, the input $x$ is pushed along the gradient direction, eventually crossing the decision boundary into the adversarial class region.}
    \label{fig:DB_ADV}
\end{figure}

A cornerstone and one of the most widely used evasion adversarial attack methods is the Fast Gradient Sign Method (FGSM) \cite{goodfellow2014explaining}. FGSM generates adversarial examples by adding perturbations in the direction of the gradient of the loss function with respect to the input. The mechanism is visualized in Figure \ref{fig:CF_ADV}, which demonstrates how an attacker climbs the loss landscape $J$ to identify a point that maximizes model error. The adversarial example is computed as:
\begin{equation}
x_{adv} = x + \epsilon \cdot \text{sign}(\nabla_{x}J(\theta,x,y))
\end{equation}
where $x$ is the original input, $x_{adv}$ is the adversarial example, $\epsilon$ is the perturbation magnitude, $\nabla_{x}J(\theta,x,y)$ is the gradient of the loss function $J$, $\theta$ represents the model parameters, and $y$ is the true label. As illustrated in Figure \ref{fig:DB_ADV}, as the epsilon parameter increases, the input $x$ is pushed along the gradient direction until it crosses the decision boundary into the adversarial class region.

\subsection{Iterative Adversarial Attacks}
While FGSM is a powerful single-step attack, more sophisticated iterative methods are often used to find more effective adversarial perturbations.

\begin{itemize}
    \item \textbf{Basic Iterative Method (BIM) \cite{kurakin2018adversarial}:} An extension of FGSM that applies the perturbation multiple times with a smaller step size $\alpha$, clipping the intermediate results after each step to ensure they stay within an $\epsilon$-neighborhood of the original input $x$.
    \item \textbf{Projected Gradient Descent (PGD): \cite{madry2017towards}} PGD is essentially a multi-step BIM with a random start point within the $\epsilon$-ball. This randomness allows PGD to explore the local loss landscape more effectively.
\end{itemize}

The relationship between model architecture and these attack trajectories raises a fundamental question: if deeper networks inherently create more convoluted decision boundaries and complex loss landscapes, does this structural complexity provide adversarial algorithms with richer pathways to exploit? A simpler, shallower boundary may prove more resilient by offering fewer gradients for both single-step and iterative algorithms to successfully calculate a label-flipping perturbation. This paper aims to investigate this hypothesis through a large-scale empirical study.

\subsection{``The Defense''}

A wide range of defenses against adversarial attacks has been proposed, including input preprocessing \cite{elshehaby2026novel}, defensive distillation \cite{papernot2016distillation}, and ensemble methods \cite{11160917}. Nevertheless, \textbf{adversarial training} is widely considered the most well-known and studied defense in the literature~\cite{bountakas2023defense}. It works by augmenting the training data with adversarial examples, exposing the model to attacks during training so that it learns to correctly classify perturbed inputs. While sometimes effective, adversarial training is computationally expensive, as it requires generating adversarial examples at every training step, and its robustness is often tied to the specific attack used during training. 

\section{Related Work}


\noindent {\bf DNN Architecture and Adversarial Robustness: }
Prior work has explored how architectural choices affect adversarial vulnerability, including the roles of activation functions~\cite{salimi2023learning}, skip connections~\cite{cazenavette2021architectural}, and model size. Huang et al.~\cite{huang2021exploring} showed that increasing model parameters
does not necessarily improve robustness, and Simon-Gabriel et al.~\cite{simon2019first} established theoretically that adversarial vulnerability scales with input dimensionality. However, none of these studies examined DNN depth in the NIDS domain, which presents fundamentally different
characteristics, lower-dimensional, structured, protocol-bound features and distinct data distributions compared to computer vision.
\noindent {\bf Depth and Adversarial Robustness in NIDS:}
ElShehaby et al.~\cite{elshehaby2025exploring} established that increasing DNN depth tends to reduce robustness in the NIDS domain, and highlighted the fundamental differences between NIDS and computer vision in the context of adversarial attacks. Our work builds directly on this finding; however, where they isolated layer depth as a single variable, we expand the investigation significantly, testing thousands of configurations across depth, feature dimensionality, activation functions, and dropout, with the goal of deriving a recipe for an \textit{inherently} robust NIDS without any explicit defense.

\noindent {\bf Hyper-Parameter Tuning for Robustness: } Mendes et al.~\cite{mendes2023hyper} explored hyper-parameter tuning to improve the effectiveness of adversarial training. While related in spirit, their work treats adversarial training as a given and optimizes around it, using computer vision datasets. Our work is positioned differently: we propose architectural design as a \textit{substitute} for adversarial training, not a companion, and we operate entirely in the NIDS domain, where the problem structure, feature space, and threat model are fundamentally distinct.

\section{Threat Model}
We assume a white-box threat model in which the attacker has full knowledge of the target model, including its architecture, parameters, and gradients. The attacker's goal is to compromise the \textit{integrity} of the NIDS by crafting adversarial network traffic that evades detection, causing malicious flows to be misclassified as benign. We further assume the attacker has access to the feature space and can manipulate input features to construct adversarial examples. We consider only gradient-based evasion attacks, specifically FGSM, PGD, and BIM, at multiple perturbation budgets~$\varepsilon$.

We acknowledge that this threat model is intentionally generous to the attacker. In practice, NIDS are internal systems that are neither exposed nor queryable by outsiders, making white-box gradient access highly unrealistic in real-world deployments~\cite{elshehaby2026evasionadversarialattacksremain}. Nevertheless, we adopt this strong threat model deliberately: if a no-defense approach proves effective even under these harsh assumptions, its practical value in more realistic settings is only stronger.
\section{Methodology and Experimental System Description}

\begin{figure}[]
\centering
\resizebox{\columnwidth}{!}{%
\begin{tikzpicture}[
  font=\tiny,
  >={Stealth},
  layer/.style={
    rectangle, rounded corners=2pt,
    minimum width=0.82cm, minimum height=0.32cm,
    draw=gray!60, thick,
    align=center, inner sep=1.5pt,
  },
  io/.style={layer, fill=gray!18},
  h64/.style={layer,  fill=blue!18},
  h128/.style={layer, fill=blue!32},
  h512/.style={layer, fill=blue!52},
  conn/.style={->, thick, gray!60},
]

\def\colsep{1.4cm}
\def\rs{0.72cm}  


\begin{scope}[xshift=0*\colsep, yshift=2*\rs]
  \node[io]  at (0,{2*\rs}) {Input ($n$)};
  \node[h64] (m1h1) at (0,{1*\rs}) {Dense 64};
  \node[io]  (m1o)  at (0,{0*\rs}) {Output ($k$)};
  \draw[conn](0,{2*\rs})--(m1h1);
  \draw[conn](m1h1)--(m1o);
  \node[font=\tiny\bfseries] at (0,{2.75*\rs}) {Model 1};
  \node[font=\tiny,gray!60]  at (0,{-0.6*\rs}) {\textit{1 hidden}};
\end{scope}

\begin{scope}[xshift=1*\colsep, yshift=1.5*\rs]
  \node[io]   at (0,{3*\rs}) {Input ($n$)};
  \node[h64]  (m2h1) at (0,{2*\rs}) {Dense 64};
  \node[h128] (m2h2) at (0,{1*\rs}) {Dense 128};
  \node[io]   (m2o)  at (0,{0*\rs}) {Output ($k$)};
  \draw[conn](0,{3*\rs})--(m2h1);
  \draw[conn](m2h1)--(m2h2);
  \draw[conn](m2h2)--(m2o);
  \node[font=\tiny\bfseries] at (0,{3.75*\rs}) {Model 2};
  \node[font=\tiny,gray!60]  at (0,{-0.6*\rs}) {\textit{2 hidden}};
\end{scope}

\begin{scope}[xshift=2*\colsep, yshift=1*\rs]
  \node[io]   at (0,{4*\rs}) {Input ($n$)};
  \node[h64]  (m3h1) at (0,{3*\rs}) {Dense 64};
  \node[h128] (m3h2) at (0,{2*\rs}) {Dense 128};
  \node[h128] (m3h3) at (0,{1*\rs}) {Dense 128};
  \node[io]   (m3o)  at (0,{0*\rs}) {Output ($k$)};
  \draw[conn](0,{4*\rs})--(m3h1);
  \draw[conn](m3h1)--(m3h2);
  \draw[conn](m3h2)--(m3h3);
  \draw[conn](m3h3)--(m3o);
  \node[font=\tiny\bfseries] at (0,{4.75*\rs}) {Model 3};
  \node[font=\tiny,gray!60]  at (0,{-0.6*\rs}) {\textit{3 hidden}};
\end{scope}

\begin{scope}[xshift=3*\colsep, yshift=0.5*\rs]
  \node[io]   at (0,{5*\rs}) {Input ($n$)};
  \node[h64]  (m4h1) at (0,{4*\rs}) {Dense 64};
  \node[h128] (m4h2) at (0,{3*\rs}) {Dense 128};
  \node[h512] (m4h3) at (0,{2*\rs}) {Dense 512};
  \node[h128] (m4h4) at (0,{1*\rs}) {Dense 128};
  \node[io]   (m4o)  at (0,{0*\rs}) {Output ($k$)};
  \draw[conn](0,{5*\rs})--(m4h1);
  \draw[conn](m4h1)--(m4h2);
  \draw[conn](m4h2)--(m4h3);
  \draw[conn](m4h3)--(m4h4);
  \draw[conn](m4h4)--(m4o);
  \node[font=\tiny\bfseries] at (0,{5.75*\rs}) {Model 4};
  \node[font=\tiny,gray!60]  at (0,{-0.6*\rs}) {\textit{4 hidden}};
\end{scope}

\begin{scope}[xshift=4*\colsep, yshift=0*\rs]
  \node[io]   at (0,{6*\rs}) {Input ($n$)};
  \node[h64]  (m5h1) at (0,{5*\rs}) {Dense 64};
  \node[h128] (m5h2) at (0,{4*\rs}) {Dense 128};
  \node[h512] (m5h3) at (0,{3*\rs}) {Dense 512};
  \node[h128] (m5h4) at (0,{2*\rs}) {Dense 128};
  \node[h64]  (m5h5) at (0,{1*\rs}) {Dense 64};
  \node[io]   (m5o)  at (0,{0*\rs}) {Output ($k$)};
  \draw[conn](0,{6*\rs})--(m5h1);
  \draw[conn](m5h1)--(m5h2);
  \draw[conn](m5h2)--(m5h3);
  \draw[conn](m5h3)--(m5h4);
  \draw[conn](m5h4)--(m5h5);
  \draw[conn](m5h5)--(m5o);
  \node[font=\tiny\bfseries] at (0,{6.75*\rs}) {Model 5};
  \node[font=\tiny,gray!60]  at (0,{-0.6*\rs}) {\textit{5 hidden}};
\end{scope}

\begin{scope}[yshift=-0.85cm]
  \node[io,   minimum width=0.55cm, label=right:{\tiny I/O}]        at (0.35, 0) {};
  \node[h64,  minimum width=0.55cm, label=right:{\tiny Dense 64}]   at (1.55, 0) {};
  \node[h128, minimum width=0.55cm, label=right:{\tiny Dense 128}]  at (3.15, 0) {};
  \node[h512, minimum width=0.55cm, label=right:{\tiny Dense 512}]  at (4.9,  0) {};
\end{scope}

\end{tikzpicture}%
}
\caption{Architectures of the five DNN models. All share the same input ($n$ features) and output ($k$ classes) layers. Hidden layer shading indicates width: light = 64, medium = 128, dark = 512 units.}
\label{fig:model_architectures}
\end{figure}
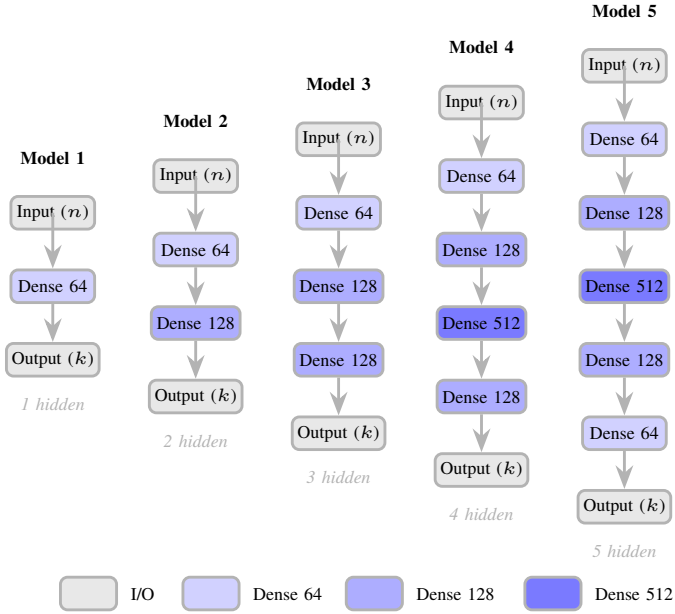

\noindent {\bf Methodology:} To explore the answer to our \hyperlink{target:RQ}{fundamental RQ}, \textit{Does the model have to be vulnerable in the first place?}, we adopt a systematic, large-scale experimental approach. If the answer is no, this immediately raises a follow-up RQ: \textit{What is the recipe for an inherently robust neural network, a concept we coin the ``no-defense defense'', against gradient-based evasion adversarial attacks in the NIDS domain?} To answer both, we evaluate approximately 2,200 model configurations, systematically varying network depth, input feature dimensionality, activation functions, and dropout across FGSM, PGD, and BIM attacks at multiple perturbation budgets. This design allows us to isolate the individual contribution of each architectural variable to adversarial robustness and observe how they interact when combined. Finally, we compare the best-performing configuration against adversarially trained deep models to evaluate whether deliberate architectural design can serve as a practical substitute for explicit defenses.



\noindent {\bf Dataset:}
We use the CSE-CIC-IDS2018 dataset~\cite{sharafaldin2018toward}, a widely adopted benchmark covering a broad range of attack categories, including Brute Force, DoS, and DDoS, alongside benign traffic. Specifically, we adopt the improved version curated by Liu et al.~\cite{liu2022error}, which addresses known issues in the original release related to erroneous feature generation and incorrect labeling, yielding a more reliable foundation for adversarial robustness research.

\noindent {\bf Pre-Processing:} To begin with, non-generalizable columns such as flow IDs and timestamps are removed, and the target variable is label-encoded. Categorical features are one-hot encoded. We also extract the geographic region from the destination IP using the \texttt{ipapi} library~\cite{ipapi} and the application layer protocol from the destination port, before one-hot encoding them. Following the caution of Arp et al.~\cite{arp2022and}, we never use raw IP addresses as numerical inputs; only derived, semantically meaningful attributes are retained to avoid spurious correlations. Ultimately, this feature engineering and encoding pipeline yields a total of 123 input features. Furthermore, numerical features are standardized to zero mean and unit variance using parameters fitted exclusively on the training set to prevent data leakage. Class imbalance is addressed via random undersampling~\cite{liu2020dealing} on the training set.

\noindent {\bf Neural Networks:} We evaluate five DNN architectures of increasing depth, ranging from one to five hidden layers. The full architectural details of each model, including layer sizes, are illustrated in Figure~\ref{fig:model_architectures}. Activation functions and dropout are intentionally varied across experiments, as the core objective of this work is to identify the architectural environment most conducive to an inherent, no-defense defense against gradient-based adversarial attacks.

\noindent {\bf Adversarial Attacks:} The Adversarial Robustness Toolbox (ART) library \cite{nicolae2018adversarial} was used to generate FGSM \cite{goodfellow2014explaining}, Projected Gradient Descent (PGD) \cite{madry2017towards}, and Basic Iterative Method (BIM) \cite{kurakin2018adversarial} adversarial attacks. 


\section{Experimental Results and Analysis}
\subsection{Baseline Performance and Feature Space Characterization}

\begin{figure}[]
    \centering
    \includegraphics[width=\linewidth]{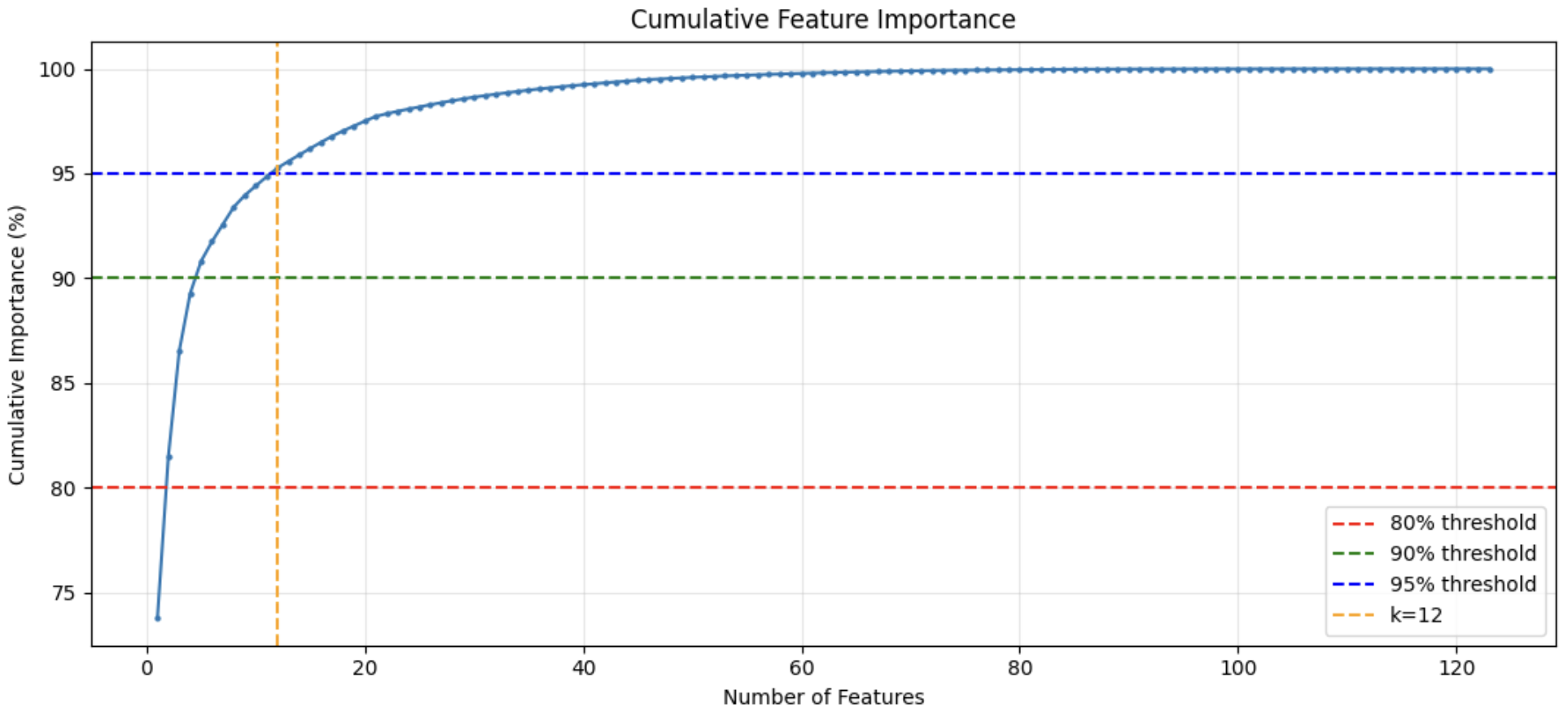}
    \caption{Cumulative feature importance across the dataset. The top 12 features account for approximately 95\% of total predictive variance.)}
    \label{fig:cumimp12}
\end{figure}

\begin{figure}[]
    \centering
    \includegraphics[width=\linewidth]{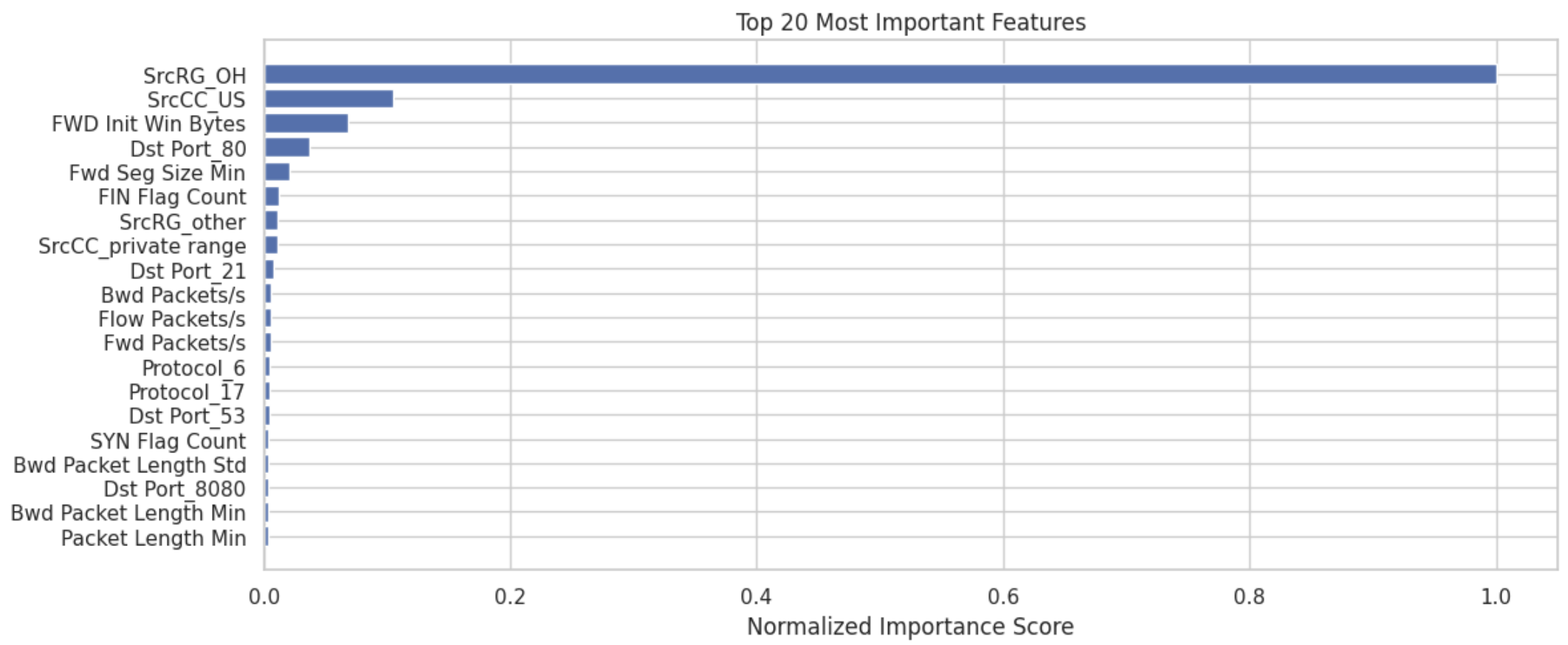}
    \caption{Normalized importance scores of the top 20 individual features, showing a steep drop-off after the highest-ranking features.}
    \label{fig:top20}
\end{figure}

To systematically evaluate the adversarial robustness of ML-based NIDS, we began by analyzing the relative importance of the dataset's features. To obtain the importance of each feature, we used the ``SelectKBest'' selector, which utilizes the Analysis of Variance (ANOVA) F-test to compute the ratio of between-group variance to within-group variance for each feature relative to the target classes. Figure \ref{fig:cumimp12} plots the resulting cumulative feature importance across the dataset. The curve demonstrates a sharp initial ascent, indicating that a very small subset of features captures the vast majority of the predictive variance. Specifically, the top 12 features account for a highly significant portion of the total importance, rapidly approaching the 95\% threshold. The disproportionate influence of these specific attributes is further detailed in Figure \ref{fig:top20}, which illustrates the normalized importance scores of the top 20 individual features. A steep drop-off in relevance is evident immediately after the highest-ranking features. This justifies our decision to establish 12 features as our minimal experimental baseline; it provides a highly constrained, yet mathematically representative, feature space to test how adversarial attacks operate when the attacker's manipulable input surface is severely limited.

\begin{table}[]

\centering

\caption{Performance of models before the attack across different feature set sizes.}
\scriptsize
\label{tab:pre_attack_performance}
\renewcommand{\arraystretch}{1}
\begin{tabular}{c l c c c c l}
\toprule
\textbf{Features} & \textbf{Model} & \textbf{Accuracy} & \textbf{Precision} & \textbf{Recall} & \textbf{F1-Score} & \textbf{T. T. (s)} \\
\midrule
\multirow{5}{*}{12}
 & Model 1 & 0.9953 & 0.9747 & 0.9465 & 0.9604 & 7.49 \\
 & Model 2 & 0.9966 & 0.9976 & 0.9465 & 0.9714 & 10.09 \\
 & Model 3 & 0.9974 & 0.9988 & 0.9576 & 0.9778 & 13.20 \\
 & Model 4 & 0.9974 & 0.9986 & 0.9582 & 0.9780 & 23.34 \\
 & Model 5 & 0.9974 & 0.9986 & 0.9582 & 0.9780 & 31.24 \\
\midrule
\multirow{5}{*}{30}
 & Model 1 & 0.9994 & 0.9910 & 0.9991 & 0.9950 & 8.05 \\
 & Model 2 & 0.9995 & 0.9924 & 0.9996 & 0.9960 & 10.90 \\
 & Model 3 & 0.9999 & 0.9989 & 0.9996 & 0.9993 & 14.04 \\
 & Model 4 & 0.9999 & 0.9981 & 0.9998 & 0.9990 & 23.54 \\
 & Model 5 & 1.0000 & 0.9998 & 0.9998 & 0.9998 & 31.23 \\
\midrule
\multirow{5}{*}{50}
 & Model 1 & 0.9994 & 0.9917 & 0.9983 & 0.9950 & 8.05 \\
 & Model 2 & 0.9996 & 0.9945 & 0.9996 & 0.9970 & 11.00 \\
 & Model 3 & 1.0000 & 0.9998 & 0.9996 & 0.9997 & 14.24 \\
 & Model 4 & 0.9999 & 0.9991 & 0.9996 & 0.9993 & 24.34 \\
 & Model 5 & 1.0000 & 0.9996 & 0.9996 & 0.9996 & 32.04 \\
\midrule
\multirow{5}{*}{75}
 & Model 1 & 0.9994 & 0.9908 & 0.9996 & 0.9952 & 8.39 \\
 & Model 2 & 0.9995 & 0.9930 & 0.9996 & 0.9963 & 11.57 \\
 & Model 3 & 1.0000 & 0.9996 & 0.9996 & 0.9996 & 17.26 \\
 & Model 4 & 1.0000 & 0.9998 & 0.9996 & 0.9997 & 25.04 \\
 & Model 5 & 1.0000 & 0.9996 & 0.9996 & 0.9996 & 32.63 \\
\midrule
\multirow{5}{*}{100}
 & Model 1 & 0.9999 & 0.9985 & 0.9996 & 0.9991 & 8.62 \\
 & Model 2 & 0.9999 & 0.9989 & 0.9998 & 0.9993 & 13.73 \\
 & Model 3 & 1.0000 & 0.9994 & 0.9998 & 0.9996 & 14.99 \\
 & Model 4 & 1.0000 & 0.9996 & 0.9998 & 0.9997 & 24.84 \\
 & Model 5 & 1.0000 & 0.9994 & 0.9998 & 0.9996 & 32.20 \\
\midrule
\multirow{5}{*}{123}
 & Model 1 & 0.9996 & 0.9934 & 0.9998 & 0.9966 & 8.95 \\
 & Model 2 & 0.9999 & 0.9994 & 0.9996 & 0.9995 & 13.24 \\
 & Model 3 & 0.9997 & 0.9956 & 0.9998 & 0.9977 & 29.79 \\
 & Model 4 & 1.0000 & 0.9996 & 0.9996 & 0.9996 & 25.62 \\
 & Model 5 & 0.9998 & 0.9961 & 0.9998 & 0.9980 & 37.06 \\
\bottomrule
\end{tabular}%
\end{table}

Having defined these subsets, we evaluated the baseline performance of all five neural network architectures on clean, unperturbed network traffic prior to any adversarial manipulation. Table \ref{tab:pre_attack_performance} details the Accuracy, Precision, Recall, F1-Score, and training times across all models and feature dimensions. The final column, denoted as T. T. (s), represents the Training Time in seconds, demonstrating the expected computational trade-off where deeper network architectures and larger feature dimensions naturally require more time to train. While the 12-feature subset achieves strong baseline accuracy (consistently above 0.995), its F1-scores, ranging from 0.9604 for the shallowest model to 0.9780 for the deepest, do not quite reach the peak performance of models trained on wider feature sets. For instance, once the feature space expands to 30 or more features, the models achieve near-perfect F1-Scores consistently exceeding 0.9950. However, the slightly lower F1-Score at 12 features is an acceptable and expected trade-off for establishing our tightly constrained baseline.

\subsection{The Effect of Neural Network Depth on Robustness}

\begin{figure}
    \centering
    \includegraphics[width=\linewidth, height=4cm]{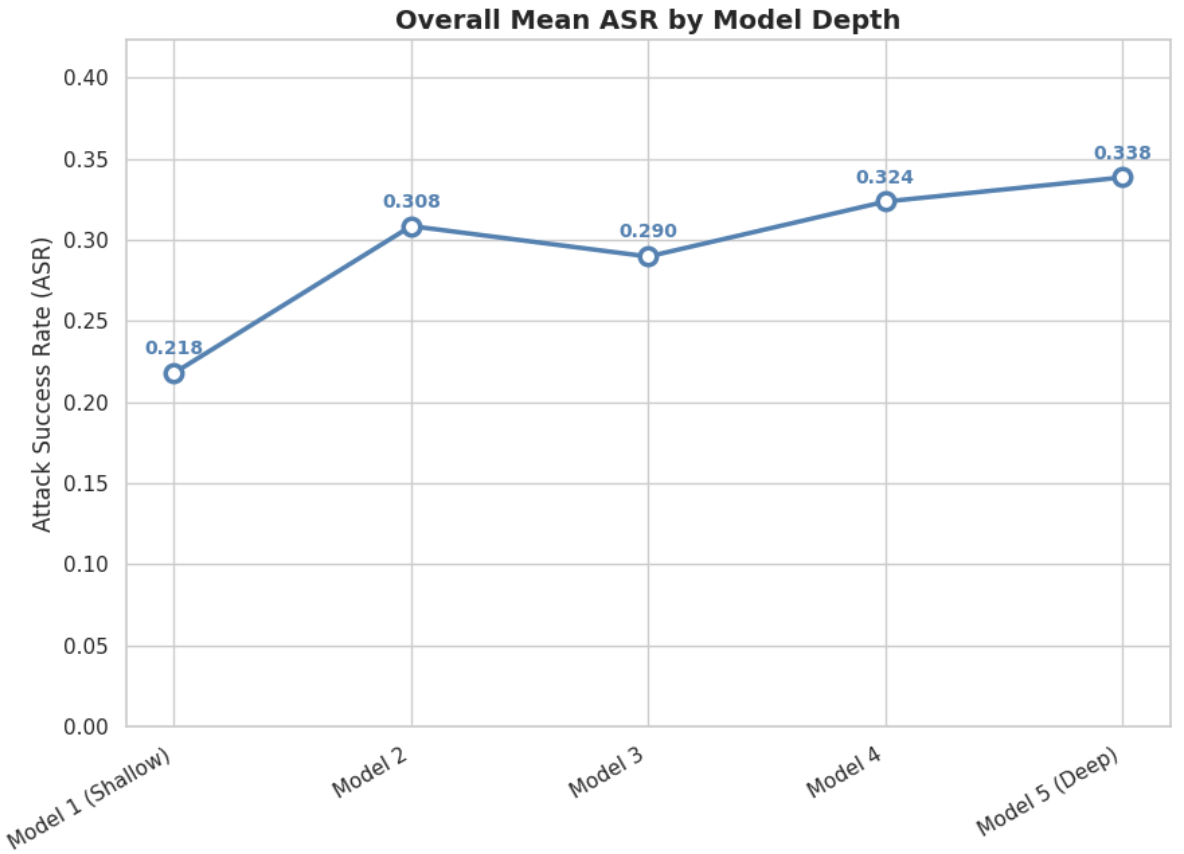}
    \caption{Mean ASR across model depths.}
    \label{fig:depthALL}
\end{figure}

\begin{figure}
    \centering
    \includegraphics[width=\linewidth, height=4cm]{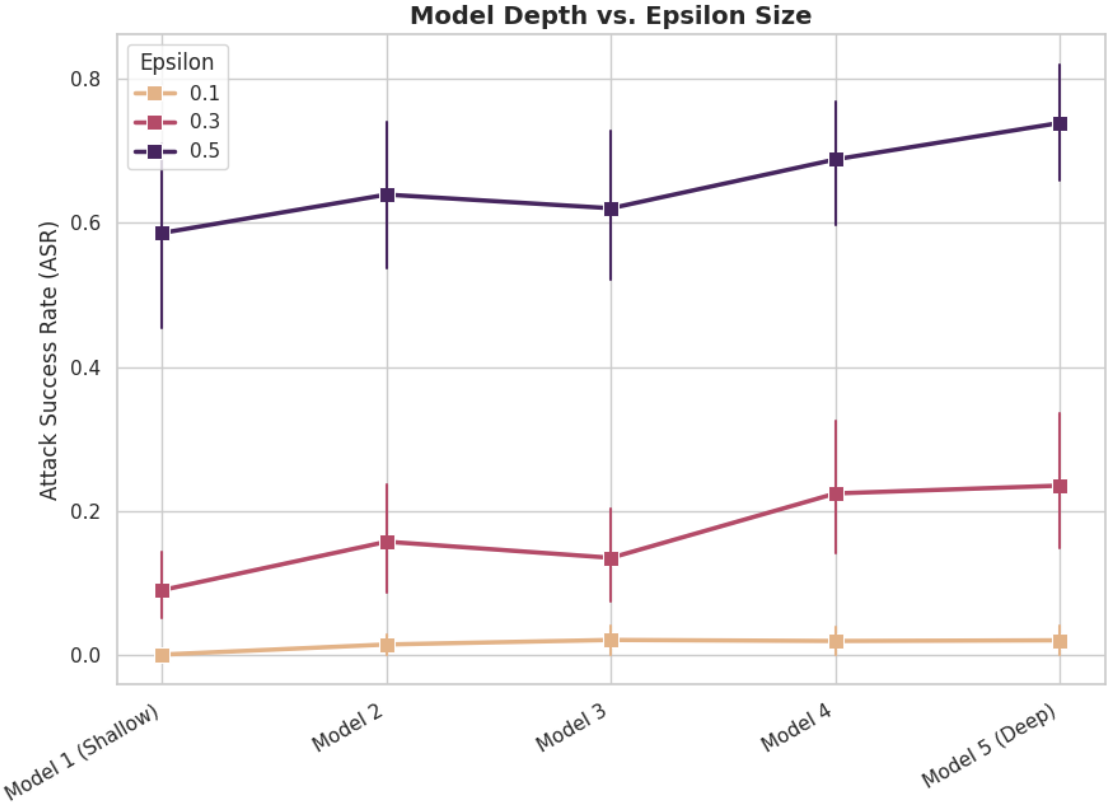}
    \caption{Effect of $\varepsilon$ on ASR across model depths.}
    \label{fig:depthvsepsilon}
\end{figure}

\begin{figure}
    \centering
    \includegraphics[width=\linewidth, height=4cm]{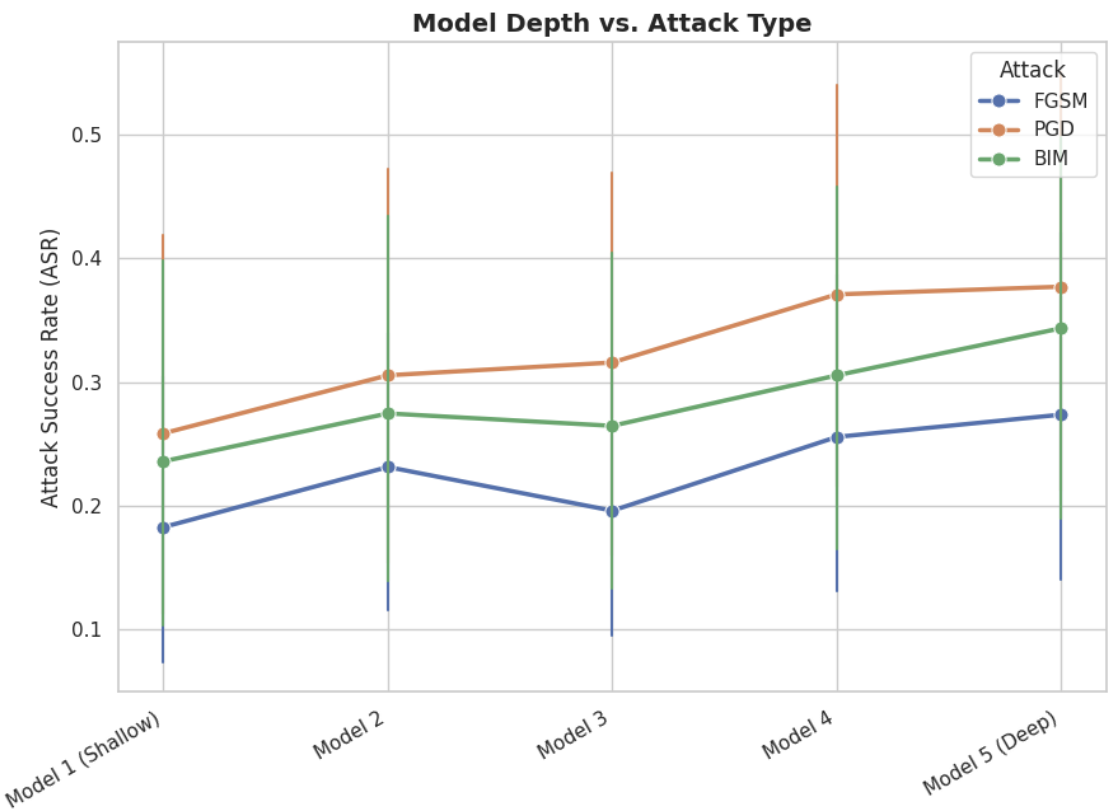}
    \caption{ASR across model depths for FGSM, PGD, and BIM.}
    \label{fig:depthvstype}
\end{figure}

\begin{figure*}[]
\centering
\includegraphics[width=\linewidth, height=8.8cm]{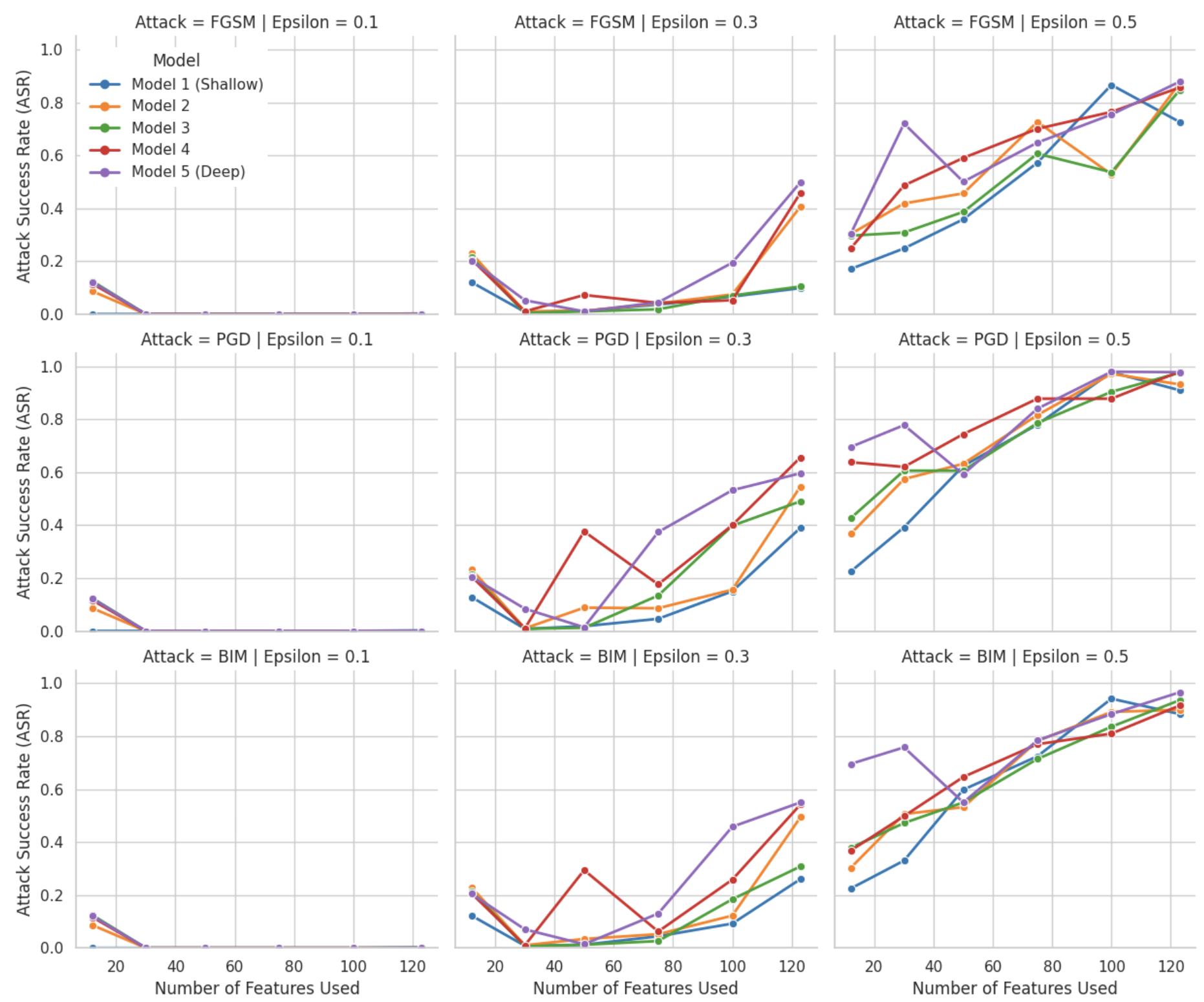}
\caption{Joint effect of feature dimensionality, model depth, and perturbation budget on ASR.}
\label{fig:all1}
\end{figure*}
With the baseline performance established, we turn to the central hypothesis of this study: \textbf{\textit{whether increasing the depth of a neural network inherently degrades its robustness against gradient-based adversarial evasion attacks in the NIDS domain.}} To measure how effectively the attacks compromise each architecture, we use the Attack Success Rate (ASR), defined as the proportion of adversarial examples that successfully caused the model to flip its prediction from the correct class to an incorrect one~\cite{elshehaby2026novel}. ASR ranges from 0 to 1, where 0 indicates no successful attacks, and 1 indicates that every crafted adversarial example fooled the model. We evaluated the ASR across the five model architectures, varying the perturbation intensity ($\epsilon$) and attack algorithms.


Figure~\ref{fig:depthALL} plots the mean ASR across all experimental conditions, aggregated over feature set sizes, perturbation budgets, and attack algorithms, for each model depth. A clear upward trend is observable: as the architecture transitions from a single hidden layer (Model 1) to five hidden layers (Model 5), the mean ASR rises consistently. This indicates that deeper networks are structurally more vulnerable on average, regardless of the specific experimental conditions. The added complexity of deeper models appears to create more convoluted decision boundaries, providing adversarial algorithms with richer gradient pathways to exploit.


This trend is also consistent across different perturbation budgets. Figure~\ref{fig:depthvsepsilon} isolates the effect of $\epsilon$ across the different model depths. As expected, a higher $\epsilon$ results in a higher ASR across all models, and deeper models remain more vulnerable regardless of the perturbation budget.


Finally, we confirm that this architectural behavior is not isolated to a single attack methodology. Figure \ref{fig:depthvstype} plots the ASR for FGSM, PGD, and BIM attacks against the varying model depths. While the more sophisticated iterative attacks (PGD and BIM) generally achieve higher success rates than the single-step FGSM, all three algorithms exhibit the exact same upward trajectory.

\subsection{Impact of Feature Dimensionality on Attack Surface}

Figure \ref{fig:all1} illustrates how expanding the input feature space (from 12 to 123 features) compounds adversarial vulnerability. A clear trend emerges: broader feature dimensionality directly increases the ASR. Interestingly, at the tightly constrained 12-feature baseline, several configurations exhibit a non-zero ASR that initially drops as the feature count increases to 30. This initial success likely occurs because with only 12 heavily weighted features available, successful perturbations inevitably alter core characteristics enough to cross the decision boundary. 

However, as the feature space expands beyond this initial dip, the ASR climbs dramatically. A wider input vector provides gradient-based algorithms with more degrees of freedom, allowing them to distribute subtle perturbations across dozens of less-critical features without violating the $\epsilon$ constraint.

\subsection{Influence of Activation Functions on ASR}

\begin{figure}[]
    \centering
    \includegraphics[width=\linewidth, height=4cm]{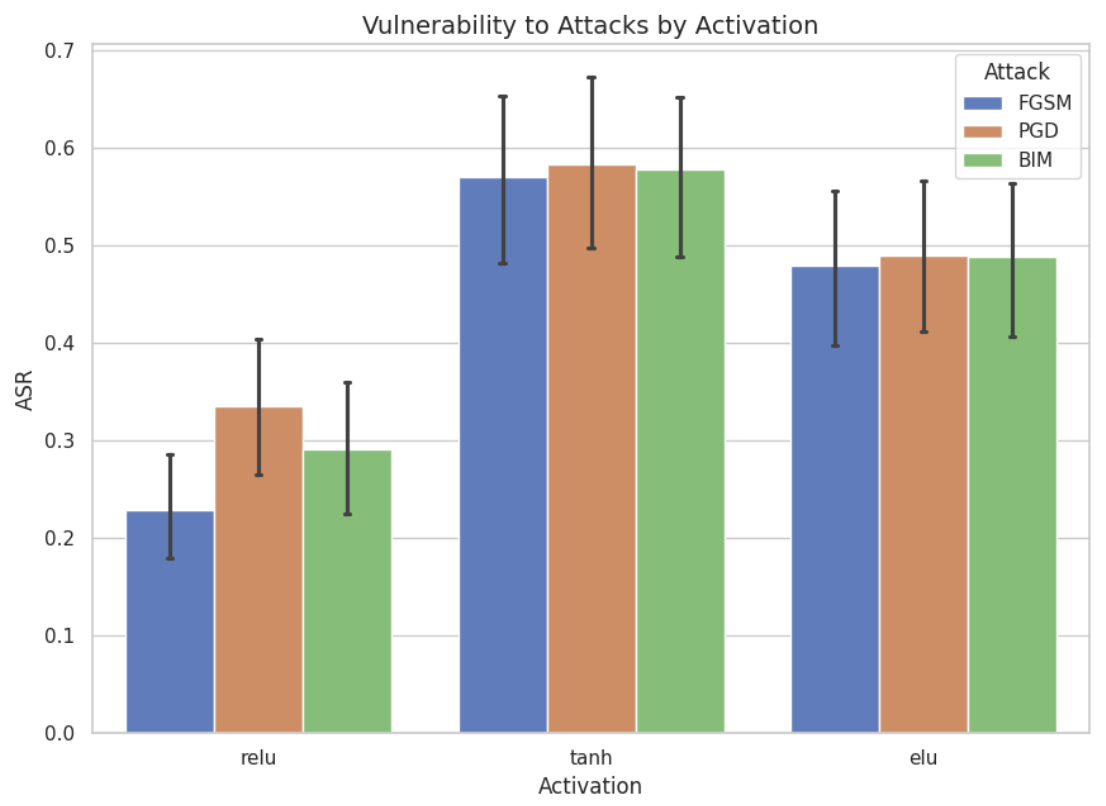}
    \caption{Mean ASR grouped by activation function across all attack types.}
    \label{fig:actfunc}
\end{figure}

\begin{figure}[]
    \centering
    \includegraphics[width=\linewidth, height=4cm]{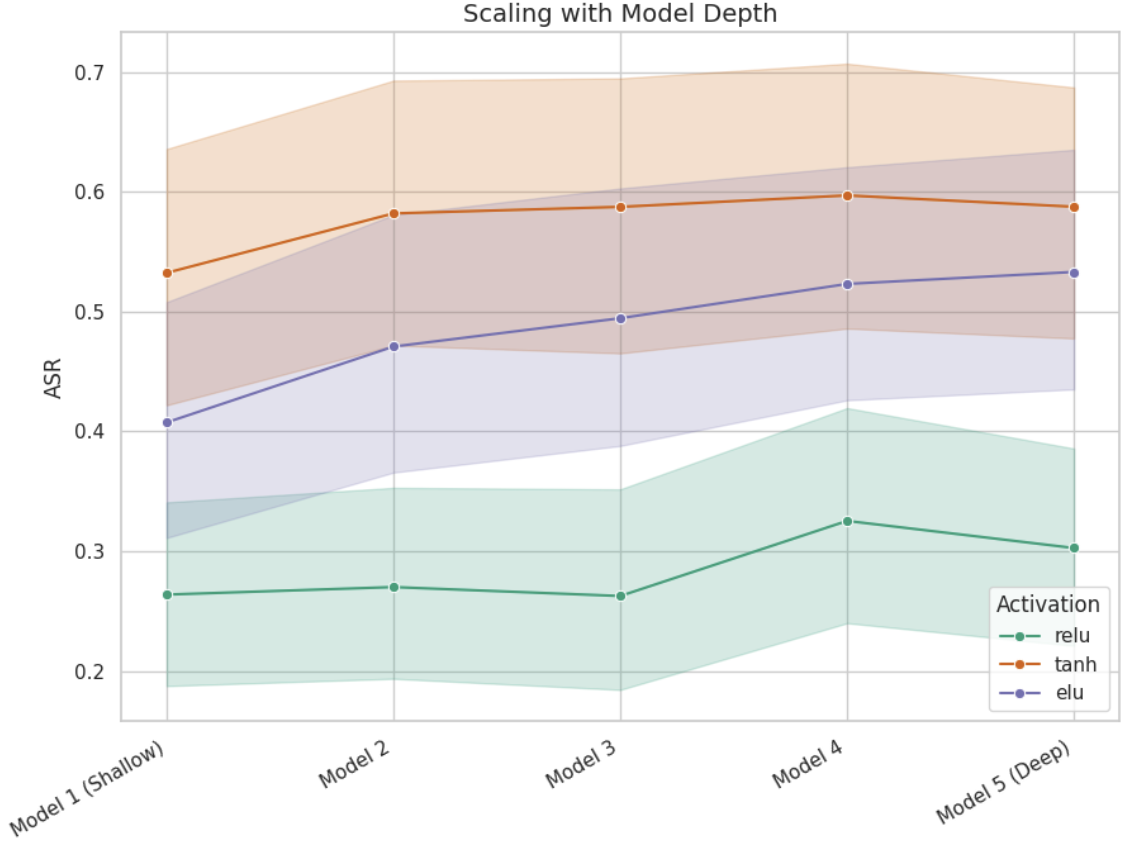}
    \caption{ASR scaling with model depth for ReLU, Tanh, and ELU.}
    \label{fig:deptgvsmodelvsact}
\end{figure}

We further investigated how the choice of activation functions affects adversarial vulnerability by evaluating the architectures using ReLU, Tanh, and ELU. Figure \ref{fig:actfunc} illustrates the overall ASR grouped by activation function across the three attack methodologies. The results clearly indicate that ReLU provides the highest robustness, maintaining a significantly lower ASR compared to the alternatives. 

This dynamic is explored across varying network architectures in Figure \ref{fig:deptgvsmodelvsact}, which plots ASR scaling with model depth for each activation function. While all three functions demonstrate the established trend of increasing vulnerability as the network deepens, their baseline vulnerability levels differ drastically. ReLU suppresses the ASR across all depths, exhibiting a much shallower degradation curve. In contrast, Tanh and ELU initiate at significantly higher vulnerability baselines even in shallow configurations and maintain this elevated ASR as depth increases. 

In Figures \ref{fig:actfunc} and \ref{fig:deptgvsmodelvsact}, the reported performance metrics represent the mean values aggregated across various experimental conditions, including differing model depths, perturbation budgets ($\epsilon$), and feature dimensions. To rigorously quantify the variance and stability of these models under different hyperparameters, our visualizations incorporate 95\% confidence intervals, represented as confidence bars in the bar charts and shaded confidence bands in the line plots. 

\subsection{Evaluating the Effect of Dropout}

\begin{figure}
    \centering
    \includegraphics[width=\linewidth, height=4cm]{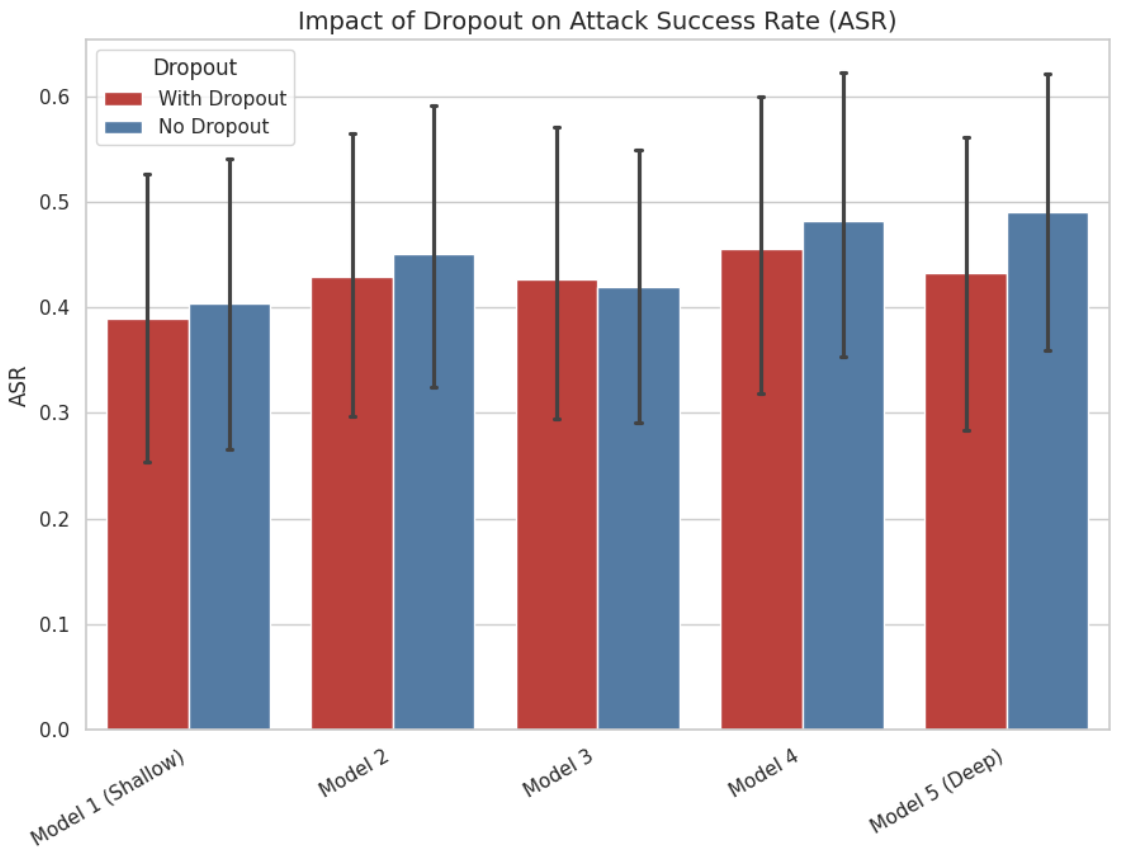}
    \caption{Effect of dropout on ASR across model architectures $(dropout$ $probability = 0.5)$}
    \label{fig:dropout}
\end{figure}

\begin{figure}[]

    \centering
    \includegraphics[width=\linewidth, height=4cm]{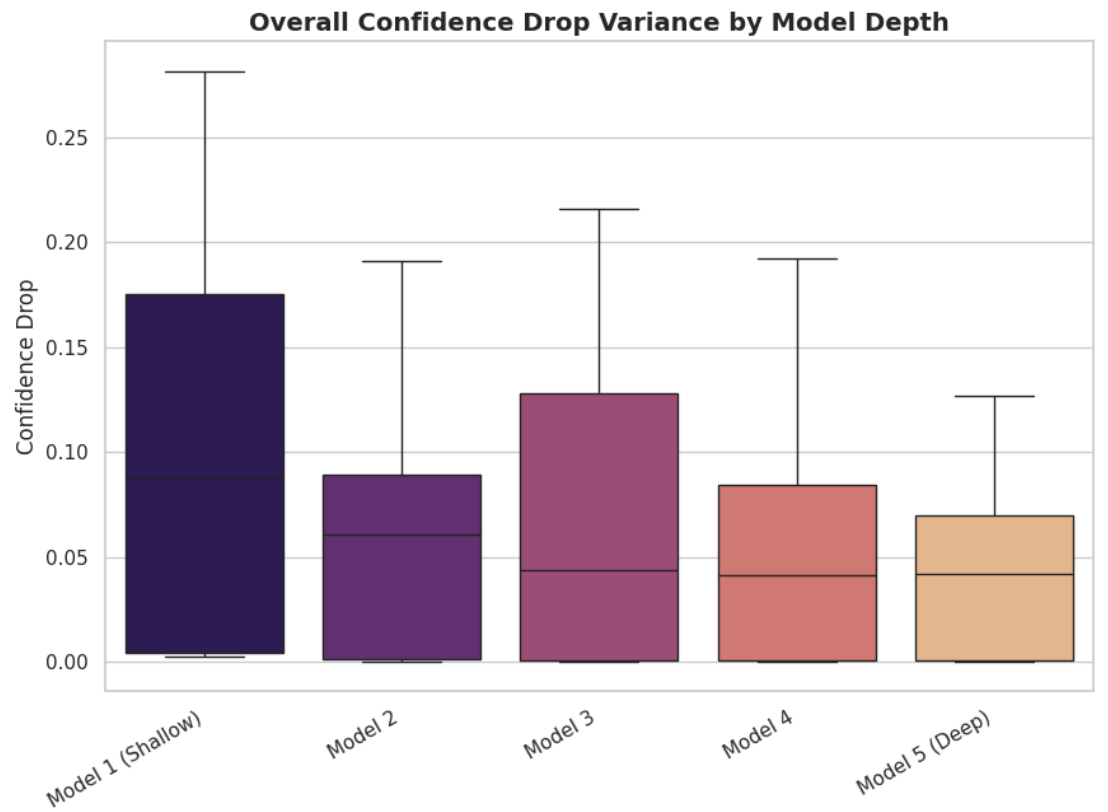}
    \caption{Distribution of confidence drop across model depths.}
    \label{fig:confvarsdepth}

\end{figure}

We further examined the impact of dropout on adversarial robustness by comparing models trained with and without a dropout probability of 0.5. As illustrated in Figure \ref{fig:dropout}, the introduction of dropout does not yield a significant shift in the ASR across the various model architectures. While dropout is traditionally employed to prevent overfitting by reducing co-dependency between neurons, our results indicate that the difference in vulnerability is marginal. In some configurations, such as Model 5, a slight decrease in ASR is noticed when dropout is applied; however, the high degree of overlap in the 95\% confidence intervals suggests that dropout is not a determinative factor in the robustness of models against gradient-based adversarial attacks.

\subsection{Confidence Drop}

\begin{figure}[]

    \centering
    \includegraphics[width=\linewidth, height=4cm]{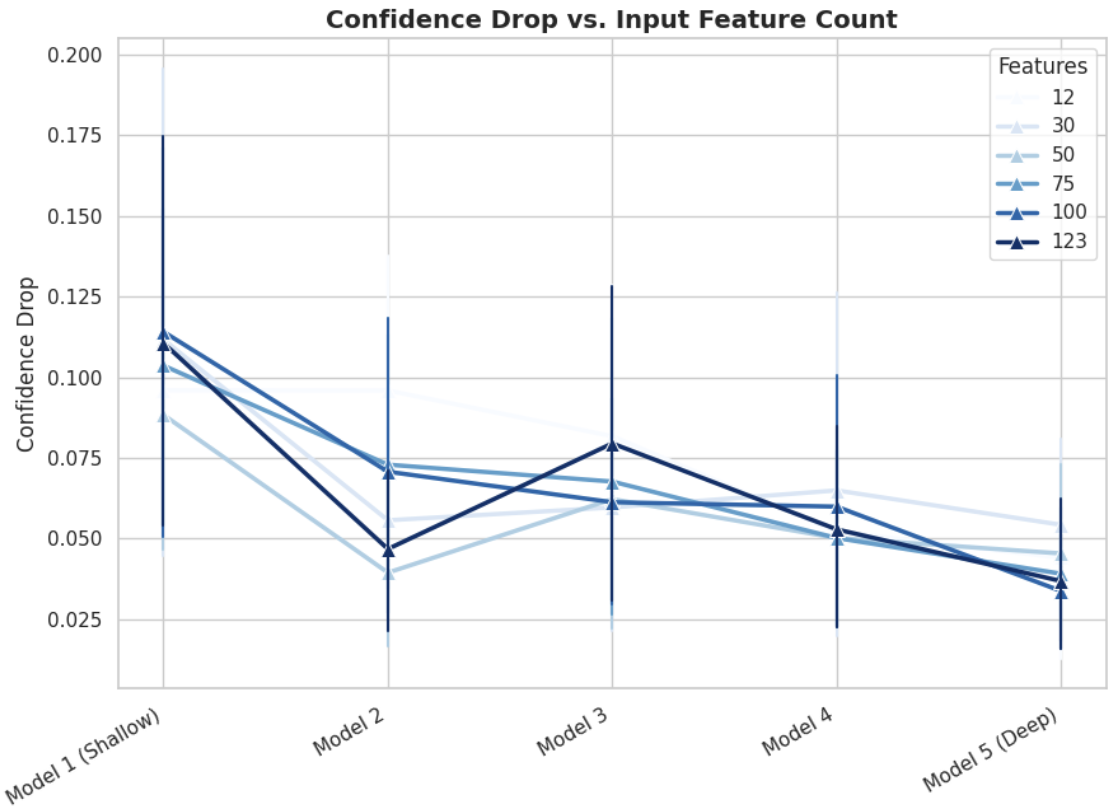}
    \caption{Confidence drop as a function of input feature count across model depths.}
    \label{fig:confvscount}

\end{figure}

\begin{figure}[]

    \centering
    \includegraphics[width=\linewidth, height=4cm]{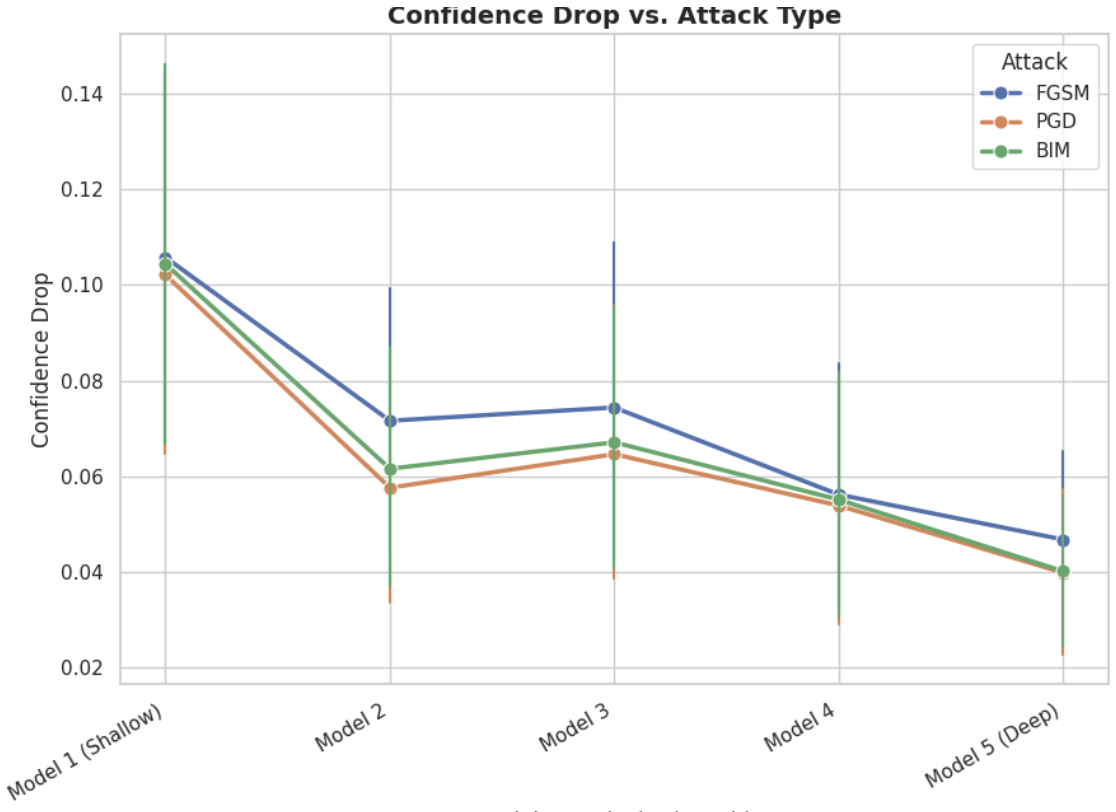}
    \caption{Confidence drop across attack types and model depths.}
    \label{fig:convstype}

\end{figure}

Analyzing the confidence reduction of the models alongside their ASR reveals a critical duality in how different architectures react to adversarial manipulation. The confidence drop is defined as the decrease in the model's predicted probability for the correct class between the original and adversarial inputs. Figure \ref{fig:confvarsdepth} shows the overall variance and magnitude of the confidence drop across model depths. The box plot shows that the shallowest network (Model 1) experiences a significant median confidence drop with a wide variance. When paired with its low ASR, this indicates that while the shallow model successfully resists label-flipping, the adversarial perturbations successfully induce a high degree of uncertainty. Conversely, the deepest network (Model 5) displays a tightly bound, minimal drop in confidence. Because deep models also exhibit a high ASR, this points to a dangerous failure mode: deeper networks easily fall for adversarial examples and confidently produce incorrect predictions. This behavior is consistent across all input dimensions. As shown in Figure \ref{fig:confvscount}, while model depth decreases the confidence drop, the drop is largely unaffected by the input feature count (12 to 123). Moreover, Figure \ref{fig:convstype} demonstrates that this trend persists regardless of the attack methodology. While iterative methods like PGD and BIM generally inflict a slightly more pronounced confidence reduction compared to the single-step FGSM, all three algorithms produce the exact same downward trajectory in confidence drop as model depth increases. Ultimately, the results confirm that attacks on deep models result in highly confident misclassifications, whereas attacks on shallow models fail to change the prediction but succeed in shaking the model's certainty.

\subsection{No-Defense Defense vs Adversarial Training}
\label{No_Defense_Defense_vs_Adversarial_Training}

Based on our preceding experimental results, we propose that an effective ``no-defense defense'' against adversarial attacks in the NIDS domain consists of designing a model with three fundamental characteristics: a shallow architecture, a constrained feature space (not overly reduced, but around 30 features depending on the dataset), and the ReLU activation function. To rigorously evaluate the robustness of this inherent structural defense, we compare our undefended baseline against the most widely used and robust defense mechanism in the field: adversarial training \cite{abou2020evaluation,abou2020investigating}. 

\begin{figure}[]

    \centering
    \includegraphics[width=\linewidth, height=4cm]{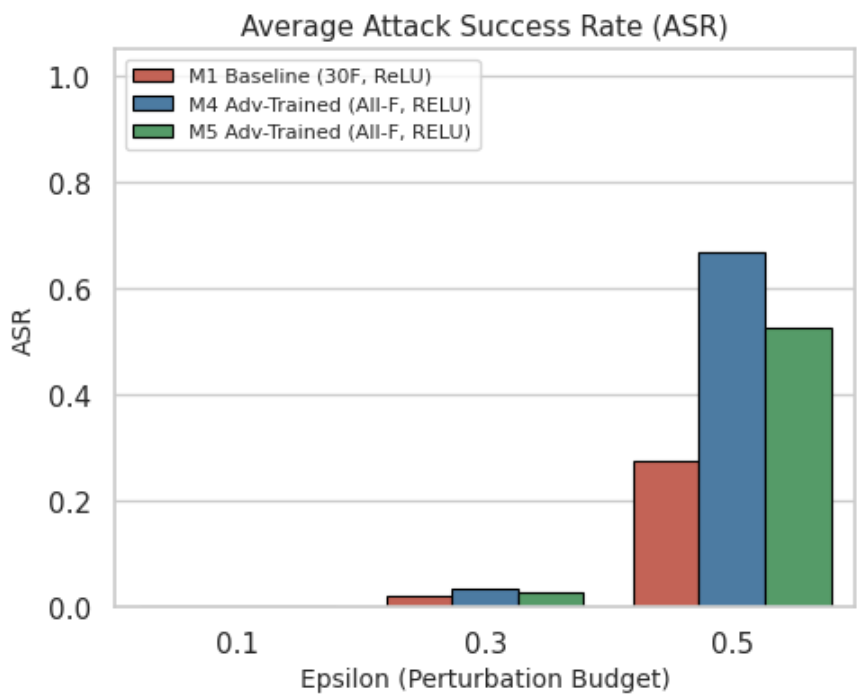}
    \caption{Evaluating the Robustness of a No-Defense Baseline Against Adversarially Trained Deep ReLU Networks}
    \label{fig:AdvTrainRELU}

\end{figure}

\begin{figure}[]

    \centering
    \includegraphics[width=\linewidth, height=4cm]{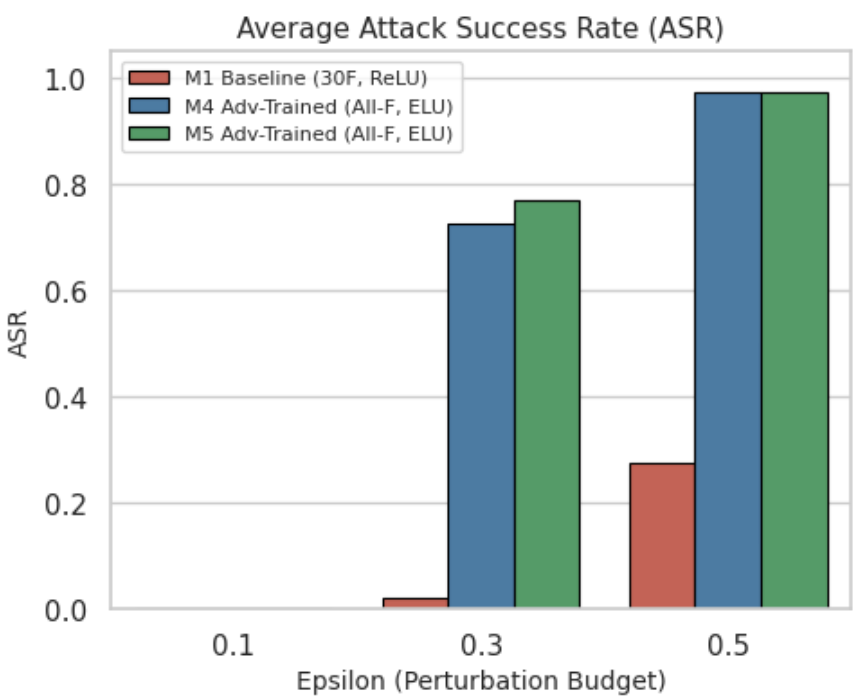}
    \caption{Evaluating the Robustness of a No-Defense Baseline Against Adversarially Trained Deep ELU Networks}

    \label{fig:AdvTrainELU}

\end{figure}

\begin{figure}[]

    \centering
    \includegraphics[width=\linewidth, height=4cm]{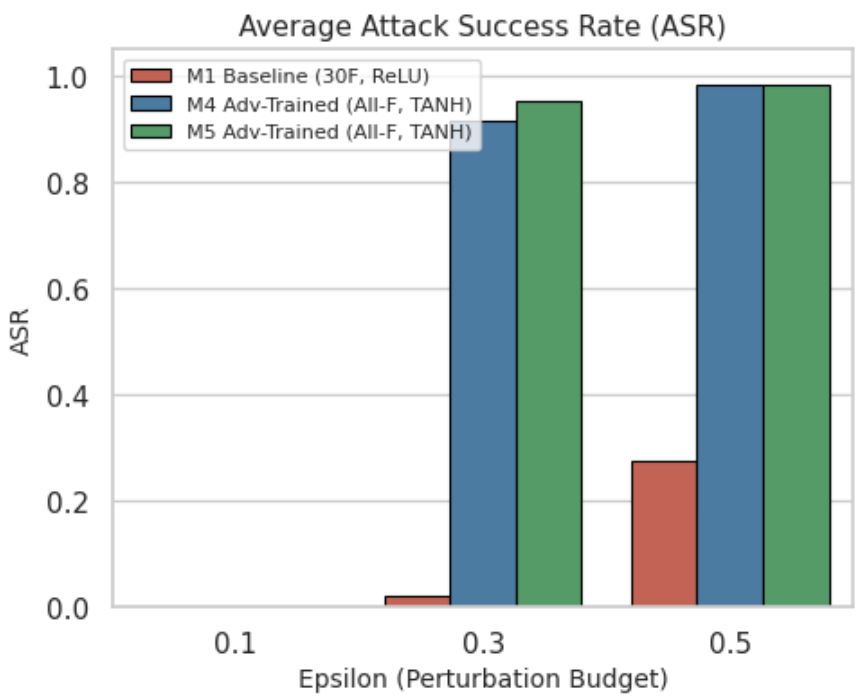}
    \caption{Evaluating the Robustness of a No-Defense Baseline Against Adversarially Trained Deep Tanh Networks}
    \label{fig:AdvTrainTanh}

\end{figure}

Figures \ref{fig:AdvTrainRELU}, \ref{fig:AdvTrainELU}, and \ref{fig:AdvTrainTanh} illustrate the Average ASR of our simple baseline (Model 1, trained on 30 features using ReLU) compared to deeper architectures (Models 4 and 5, trained on all features) that have been adversarially trained across ReLU, ELU, and Tanh activation functions, respectively.

The data reveal that the structural resilience of the optimal shallow model consistently outperforms the post-hoc defense applied to deeper networks. As demonstrated in Figure \ref{fig:AdvTrainRELU}, at a severe perturbation budget of $\epsilon = 0.5$, the undefended shallow baseline successfully restricts the ASR to below 0.3. Conversely, the deeper models, despite being adversarially trained, suffer ASR spikes exceeding 0.5 and 0.6.


This disparity in robustness is even more drastic when examining saturating activation functions. Figures \ref{fig:AdvTrainELU} and \ref{fig:AdvTrainTanh} show that adversarially trained deep models utilizing ELU and Tanh suffer near-total compromise (ASR approaching 1.0) under high perturbation ($\epsilon = 0.5$). In  contrast, our undefended baseline maintains its comparatively low vulnerability. 

Ultimately, these comparisons demonstrate that architectural simplicity provides a superior, naturally robust foundation. In NIDS applications, strategically limiting model depth and feature dimensions while utilizing ReLU yields better adversarial resistance than applying computationally expensive adversarial training to inherently vulnerable deep networks.

\section{Discussion}

\subsection{Is less more?}

The short answer is \textbf{\textit{yes, but not a blind yes}}. Across experiments, reducing architectural complexity and input dimensionality consistently improves
robustness against gradient-based adversarial attacks. However, the relationship is not monotonic, and blindly minimizing every design dimension can backfire. 

\textbf{For layer depth}, the trend is clear. As seen in Figure \ref{fig:depthvstype}, regardless of the underlying mathematical approach used to craft adversarial examples, whether the single-step FGSM or the iterative PGD and BIM, increasing the number of hidden layers expands the usable attack surface. As the architecture grows from one hidden layer (Model~1) to five (Model~5), the ASR generally rises across almost all feature set sizes, perturbation budgets, and attack types. Deeper networks usually create more convoluted decision boundaries, giving gradient-based algorithms richer pathways to exploit. Consequently, for NIDS applications where the feature space is rigidly structured by network protocol constraints, the \textit{deeper is better} paradigm is a dangerous misconception; shallow networks provide a strong, natural, no-defense defense against gradient-based manipulations. \textbf{Using fewer features} similarly reduces the attack surface, but the picture is more nuanced. The general trend holds, a wider input vector gives attackers more
degrees of freedom to distribute perturbations, but as seen in Fig.~\ref{fig:all1}, at $\varepsilon = 0.3$ and $\varepsilon = 0.1$, the 12-feature models actually show a \textit{higher} ASR than the 30-feature
models, before the expected upward trend resumes. With only 12 heavily weighted features, even a small perturbation inevitably disturbs the most discriminative
inputs, making it easier to cross the decision boundary despite the tight surface. Furthermore, as shown in Table~\ref{tab:pre_attack_performance}, the
12-feature models carry a clean performance cost, with F1-Scores falling noticeably below those of models trained on 30 or more features. So the minimum
is not the optimum.

These results highlight that depth, feature count, and activation function are correlated, interconnected design knobs. The evidence strongly favors smaller models, but the sweet spot lies in deliberate, informed reduction, not blind minimization. We should certainly \textit{explore less}, but always with the full picture in view.


\subsection{Do we need all these features?}
A clear trend is emerging in the NIDS research community: datasets are growing wider with every generation. Before one-hot encoding, the UNSW-NB15, 2015, dataset contained features in the 40s, the CSE-CIC-IDS, 2018, dataset expanded into the 80s, and the more recent BCCC-CSE-CIC-IDS2018, 2025, dataset now carries over 300 features. The implicit assumption driving this trend is that more features mean better detection; nevertheless, our results challenge that assumption directly.

ElShehaby et al.~\cite{elshehaby2026novel} demonstrated that a carefully selected subset of features is sufficient to build outstanding detection models, and our experimental results strongly agree. As shown in Table~\ref{tab:pre_attack_performance}, models trained on as few as 30 features achieve near-perfect F1-Scores, matching or exceeding the performance of models trained on the full feature set, while also being significantly faster to train. This is consistent with the broader observation by Grosse et al.~\cite{grosse2024towards}, who noted that datasets in AI security research tend to carry far more features than are encountered in real-world deployments. From an adversarial robustness perspective, feature bloat is not merely inefficient; it is a liability. As demonstrated in Fig.~\ref{fig:all1}, a wider input vector directly expands the attack surface, giving gradient-based algorithms more degrees of freedom to distribute perturbations across features without triggering detection. In short, the additional features slow down training, add no meaningful detection gain, and make the model measurably easier to attack. We do not need all these features.

\subsection{Why ReLU?}

\begin{figure}[h]
    \centering
    \begin{tikzpicture}
        \begin{axis}[
            width=\linewidth,      
            height=5.5cm,          
            xmin=-4, xmax=4,
            ymin=-0.2, ymax=1.3,   
            axis lines=middle,
            xlabel={$x$},
            ylabel={$f'(x)$},
            xlabel style={anchor=west},
            ylabel style={anchor=south},
            grid=both,
            grid style={dashed, gray!30},
            tick label style={font=\scriptsize},
            legend style={
                at={(0.95,0.5)}, 
                anchor=east,
                font=\scriptsize,
                draw=gray!50,
                fill=white,
                cells={align=left}
            },
            every axis plot/.append style={thick}
        ]

        \addplot[blue, domain=-4:0, samples=2] {0};
        \addlegendentry{ReLU'}
        \addplot[blue, domain=0:4, samples=2, forget plot] {1};

        \addplot[red, densely dashed, domain=-4:4, samples=100] {1 - (tanh(x))^2};
        \addlegendentry{tanh'}

        \addplot[green!60!black, dashdotted, domain=-4:4, samples=100] {x<=0 ? exp(x) : 1};
        \addlegendentry{ELU' ($\alpha=1$)}

        \end{axis}
    \end{tikzpicture}
    \caption{Derivatives of the ReLU, Tanh, and ELU activation functions. ReLU (blue) outputs exactly zero for any negative input, while Tanh (red, dashed) and ELU (green, dash-dotted) maintain non-zero derivatives throughout.}
    
    \label{fig:activation_derivatives}
\end{figure}
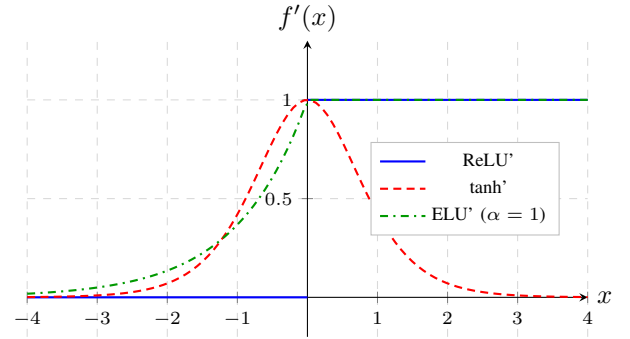

The superior robustness of the ReLU activation function stems fundamentally from its piecewise linear mathematical structure, which naturally induces a defensive phenomenon known as gradient masking. As seen in Figure \ref{fig:activation_derivatives}, unlike smooth, saturating functions like Tanh and ELU, which maintain continuous, non-zero derivatives that  attack algorithms can easily trace backward to calculate optimal input perturbations, ReLU outputs an exact zero derivative for any negative input. During an attack, these zero-state neurons effectively shatter the backward propagation of the loss gradient, reducing the adversarial algorithm's ability to identify perturbation pathways. Furthermore, this property creates inherent network sparsity, which severely restricts the manipulable attack surface within the already tightly constrained NIDS feature space, making it significantly harder for attackers to mathematically map and exploit the model's decision boundaries.


\subsection{Are these attacks practical in the NIDS domain?}

Gradient-based attacks are a powerful research tool; however, they are not particularly practical in the real-world NIDS domain. A real attacker does not have white-box access to a deployed model's gradients, and NIDS systems are not designed to be queried by outsiders. Furthermore, NIDS features are highly correlated and constrained, making these attacks even more infeasible. Apruzzese et al.~\cite{apruzzese2022position} noted that there is no concrete evidence that any real-world attacker has ever used gradients to conduct an intrusion evasion attack. Moreover, the authors argue that the most practical attacks are simple, gradient-free blind perturbations~\cite{apruzzese2024adversarial} that require no model knowledge, no query access, and are trivially easy to generate~\cite{elshehaby2026evasionadversarialattacksremain}. This context makes our no-defense defense even more compelling, as most defenses against gradient-based adversarial perturbations are expensive and provide high overhead with low return in the NIDS domain.



\subsection{The Win-Win Argument of a No-Defense}

The findings of this paper converge on a result that is, at first glance, counterintuitive: doing \textit{less} yields \textit{more} on every front simultaneously. As demonstrated in Section~\ref{No_Defense_Defense_vs_Adversarial_Training}, our proposed recipe, a shallow architecture, approximately 30 features (for this dataset), and ReLU activation, consistently outperforms deeper, all-feature models that were adversarially trained, the most popular and effective yet computationally expensive defense in the field, while simultaneously achieving near-perfect F1-Scores on clean traffic and significantly lower training times. There is no defense to implement, no overhead to absorb, and no performance penalty to accept. The architecture itself \textit{is} the defense, not a perfect one: although it significantly reduces the ASR in most cases, it does not nullify the attacks, though neither do most explicit defenses. It is, however, a strong, practical, and inherently built-in layer of resistance that comes at no additional cost.


\section{Conclusion}

This paper demonstrated that deliberate architectural choices can serve as an inherent implicit defense against gradient-based adversarial attacks on DNN-based NIDS, without any explicit defense mechanism. Across thousands of experiments, we consistently found that shallower networks, reduced feature dimensionality, and ReLU activation each reduce adversarial vulnerability, and that their effects
compound when combined. Dropout, by contrast, had a less drastic impact on robustness. A model following this recipe, one hidden layer, approximately 30 features, and ReLU, not only outperformed adversarially trained deep models in robustness, but also achieved near-perfect clean-traffic detection at a fraction of the training cost. The architecture itself is the defense. Future work should explore the generalizability of these findings across other NIDS datasets and investigate the interaction between this
inherent robustness and practical, domain-constrained adversarial perturbations.

\bibliographystyle{IEEEtran}
\bibliography{main}

\end{document}